\documentclass{article}

\usepackage{microtype}
\usepackage{booktabs} 

\usepackage{pgfplots}
\pgfplotsset{compat=newest}

\usepackage{xcolor}
\usepackage{tikz}
\definecolor{1}{HTML}{990000}
\definecolor{2}{HTML}{003366}
\definecolor{3}{HTML}{004080}
\definecolor{4}{HTML}{004d99}

\usepackage{color}
\definecolor{darkgreen}{rgb}{0,0.6,0.2}

\usepackage{graphicx} 
\usepackage{caption}
\usepackage{subcaption}
\usepackage{wrapfig}

\usepackage[accepted]{icml2018}

\icmltitlerunning{Geometric Generalization Based Zero-Shot Learning}

\begin{document}

\twocolumn[
\icmltitle{Geometric Generalization Based Zero-Shot Learning\\ Dataset Infinite World: Simple Yet Powerful}




\begin{icmlauthorlist}
\icmlauthor{Rajesh Chidambaram}{tu,cmu1}
\icmlauthor{Michael Kampffmeyer}{uit,cmu1}
\icmlauthor{Willie Neiswanger}{cmu}
\icmlauthor{Xiaodan Liang}{cmu} \\
\icmlauthor{Thomas Lachmann}{tu,bel}
\icmlauthor{Eric Xing}{cmu,pet}

\end{icmlauthorlist}

\icmlaffiliation{tu}{Center for Cognitive Science, TU Kaiserslautern, Germany}
\icmlaffiliation{bel}{Department of Experimental Psychology, University of Leuven, Leuven, Belgium}
\icmlaffiliation{uit}{UiT Machine Learning Group, UiT The Arctic University of Norway, Troms{{\o}}, Norway}
\icmlaffiliation{cmu}{Machine Learning Department, Carnegie Mellon University, USA}
\icmlaffiliation{cmu1}{Visiting Scholar, Carnegie Mellon University, USA}
\icmlaffiliation{pet}{Petuum Inc, USA}

\icmlcorrespondingauthor{Rajesh Chidambaram}{rajesh1990@live.in}

\icmlkeywords{generative models, zero-shot learning, generalization, spatial reasoning, numerical reasoning, human-level intelligence, optimization}

\vskip 0.3in
]



\printAffiliationsAndNotice{}  

\begin{abstract}
Raven's Progressive Matrices are one of the widely used tests in evaluating the human test taker's fluid intelligence. Analogously, this paper introduces geometric generalization based zero-shot learning tests to measure the rapid learning ability and the internal consistency of deep generative models. Our empirical research analysis on state-of-the-art generative models discern their ability to generalize concepts across classes. In the process, we introduce \textit{Infinite World}\footnote{The dataset will be open-sourced on GitHub}, an evaluable, scalable, multi-modal, light-weight dataset and Zero-Shot Intelligence Metric ZSI. The proposed tests condenses human-level spatial and numerical reasoning tasks to its simplistic geometric forms. The dataset is scalable to a theoretical limit of infinity, in numerical features of the generated geometric figures, image size and in quantity. We systematically analyze state-of-the-art model's internal consistency, identify their bottlenecks and propose a pro-active optimization method for few-shot and zero-shot learning.
\end{abstract}
\section{Introduction}

A remarkable case of zero-shot learning by the human brain can be drawn from one of the greatest minds of the 20th century, Albert Einstein. On the basis of his general theory of relativity, Einstein predicted gravitational waves \citep{einstein1937gravitational}, decades before the actual experiments confirmed its existence \citep{abbott2016observation}. Like no other modern physicist, his discoveries fundamentally altered and expanded our understanding of nature \citep{folsing1997albert}. The ability to perform description based learning and simulation based planning, are vital aspects of human intelligence. Unlike any other species, human brains can rapidly acquire knowledge through natural language conversations and by reading books. The human brain comprehends textual content by constructing mental models of the text \citep{woolley2011reading}. It is hypothesized that the brain optimizes cost functions in a diverse, region-specific and developmental stage dependent manner \citep{marblestone2016toward}.

On the other hand, Machine Learning (ML) methods such as Artificial Neural Network (ANN), powered by deep learning techniques \citep{lecun2015deep,schmidhuber2015deep} have proven to be powerful function approximators. Their applications span across image classification, audio processing, game playing, machine translation, etc., \citep{sprechmann2018memory}. Despite their success, their ability to zero-shot learn via simulation  \citep{ha2018world} and perform human level rapid learning from mere textual descriptions \citep{chaplot2017gated} are experimental. Specifically, when the data availability is scarce, simulation based few-shot learning methods becomes indispensable. 

In this paper, we introduce human-level zero-shot tests for text-to-image synthesis models and a Zero-Shot Intelligence Metric ZSI. The standardized tests posed by the dataset \textit{Infinite World}, consists of human-level 2-dimensional geometric generalization tasks.

\begin{figure*}[t!]
  \centering
  \includegraphics[width=\textwidth]{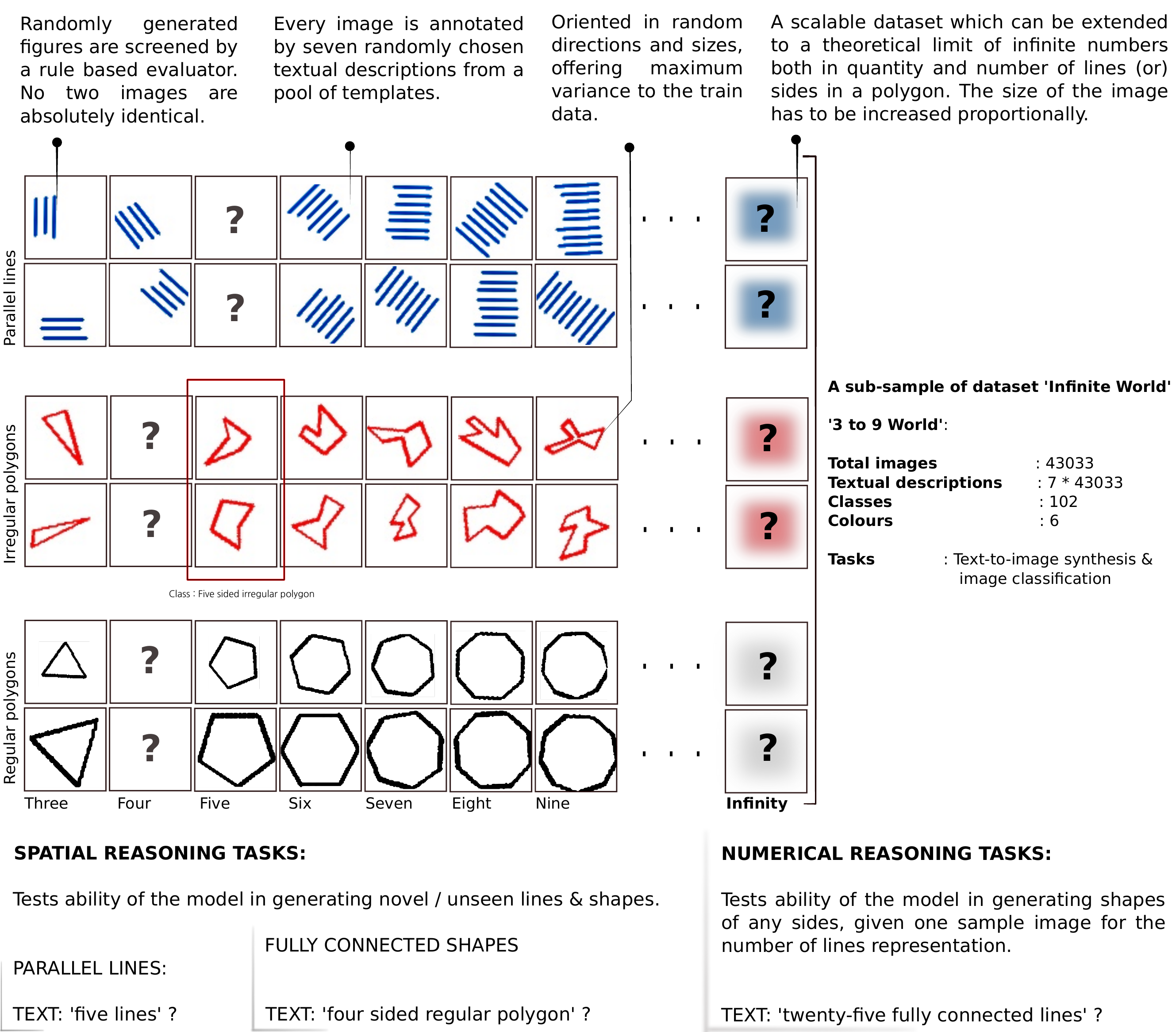}
    \caption{Zero-shot learning tests on 2-dimensional geometric generalizations tasks}
\end{figure*}

\subsection{Generative models for text-to-image synthesis}

The advent of Generative Adversarial Networks (GANs) \citep{goodfellow2014generative} in generating realistic images, image restoration, video generation, text-to-image synthesis \citep{radford2015unsupervised,zhang2016stackgan,xu2017attngan} have been impressive. Upon training, these models are able to generate images of faces, birds and room interiors from mere textual inputs. Yet, the ability of such models to generate images on unseen texts that involve human-level reasoning and rapid learning are experimental. In this paper, we systematically analyze the performances of state-of-the-art text-to-image synthesis models on the proposed 2d-geometric generalization tasks. Through the Zero-Shot Intelligence metric ZSI ($\psi$) a precise analysis of such text-to-image synthesis models reveal their internal consistency and caveats in image generation.

\subsection{Zero-Shot Learning}

The demand for vast supervised data has made deep neural networks incompatible with several cardinal tasks. Inspired by the rapidly learning nature of human-beings, zero-shot learning aims to imitate such behaviors in machines. Few-shot learning and zero-shot learning are extreme forms of transfer learning \citep{goodfellow2016deep}. Zero-shot learning is also called as zero-data learning. Popular few-shot adaptation techniques include bayesian modeling \citep{tenenbaum1999bayesian}, fine tuning a pre-trained neural net \citep{yu2010roles}, memory augmented networks \citep{santoro2016one} and meta-learners \citep{schmid1987,ravi2016optimization}. In this paper, we shall see the necessity of more advanced optimization techniques for generative models, to perform zero-shot learning on human-level reasoning tasks.

\subsection{Geometric generalization}

A three-sided triangle, a four-sided quadrilateral, and so on to $n$ sides can be generalized to the geometric concept of \textit{polygon}. It is sufficient to teach a human the concept of a polygon for few numbers and yet the person can generalize the concept to unseen numbers. Such a generalization is possible for any $n$ dimensional space ($n>1$). Whereas, state-of-the-art machine learning methods demands examples from every new class.  In this work, we approach 2d-geometric generalization tasks with ANNs. Utilizing ANNs for such tasks leverages the ability of machine learning algorithms to perform learning with comparatively less human input. Upon developing a domain specific model which can perform zero-shot learning on complicated tasks, the model can be scaled to real world applications by expanding its verbal and visual corpus.

\section{Dataset: \textit{Infinite World}}

In general, existing benchmarks such as the visual genome \cite{krishna2017visual} and the CLEVR dataset \cite{johnson2017clevr} proposes to test the reasoning abilities of the machine learning algorithms. Models such as the relational network \cite{santoro2017simple}, which successfully solves visual question answering (VQA) problems on the CLEVR datasets, suffers severe limitations \cite{kim2018notsoclevr}. Alternatively, it is notable that images from physics and geometry were used to demonstrate \textit{why a diagram is sometimes worth ten thousand words} \citep{larkin1987diagram}. On a similar note, we propose geometric generalization based zero-shot learning tests for methods in ML. The proposed tests condenses human-level reasoning tasks to its simplistic geometric forms. This simplicity aids in precisely evaluating the performance of the generative model. The dataset is scalable to a theoretical limit of infinity, in numerical features of the generated geometric figures, image size and in quantity. It is notable that the newly introduced dataset \textit{Infinite World} and Zero-Shot Intelligence metric are not restricted to text-to-image synthesis algorithms, but also for other machine learning tasks such as classification.

\textit{Infinite World} proposes tasks that are comparable to fluid intelligence tests such as Cattell Culture Fair IQ test \citep{cattell1963theory} and Progressive Matrices \citep{duncan1995fluid}. While deducing the intelligence of an agent to a single metric is difficult, we formulate a task specific zero-shot intelligence metric $\psi$ to compare the zero-shot performances of different models. Metric $\psi$ is always accompanied by the task in consideration. This offers ANN models a common platform to test their zero-shot performances on tasks including human level spatial and numerical reasoning. The generated images are verified for its accuracy by a rule-based evaluator. For every rejected generation, a new figure is generated for the required textual description of the image. The same rule-based evaluator is later used to evaluate the performance of the text-to-image synthesis model.

\subsection{\textit{3-9 World}: A subsample of \textit{Infinite World}}

The \textit{3-9 World} dataset, a subset of the \textit{Infinite World} dataset generator, considers that only numbers from $3\ to\ 9$ exists.  The sub-sampled dataset consists of $41,000\ +$ train images of lines and polygons. The images are of size $64 \times 64$ pixels. Each image consists of seven randomly-chosen textual descriptions from a pool of templates. The dataset consists of 567 distinct train texts and 91 distinct zero-shot test texts. The pool of textual annotations are customizable and scalable. The model is required to generalize the concept of connectedness of shapes and non-connectedness of lines; the zero-shot task is to predict a particular \textit{unseen} shape from the generalized concept. From our preliminary experiments, we observed that the generative models were comparatively better in generating images of same color as in the text. But they developed significant gap in performance while generating disconnected lines and connected shapes. Hence, the dataset was further developed to mainly focus on parallel lines, irregular polygons and regular polygons.

\section{Zero-Shot Intelligence metric ZSI ($\psi$)}

Popular text-to-image synthesis models are tasked for zero-shot generation of realistic images of bedrooms, flowers and human faces. In such tasks, it is difficult to precisely evaluate the ability to reason, zero-shot learn and infer consistency. Existing evaluation metrics such as the inception score \citep{salimans2016improved} and R precision \citep{xu2017attngan} do not capture the generative abilities of the model \citep{barratt2018note}. Unlike real world images, since the \textit{Infinite World} condenses complex reasoning tasks to simplistic geometric forms, it is possible to precisely evaluate the model's performance on a zero-shot metric. Subjective nature of evaluation is comparatively minimal in such tasks. Hence, we introduce a new evaluation method for Zero-Shot Intelligence ZSI ($\psi$) to measure the reliability of black-box function approximators.

\begin{figure}[h!]
  \centering
  \includegraphics[width=0.45\textwidth]{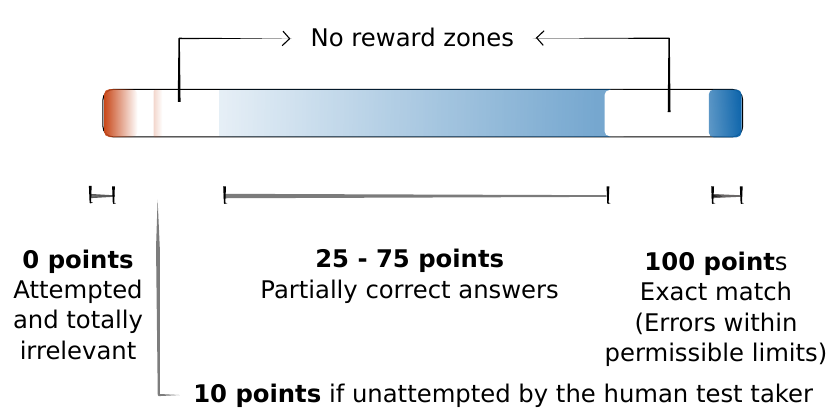}
    \caption{ZSI ($\psi$) - scoring method}
\end{figure}

\begin{figure*}[t!]
    \centering
    \begin{subfigure}[b]{0.5\textwidth}
        \centering
        \includegraphics[height=1.8in]{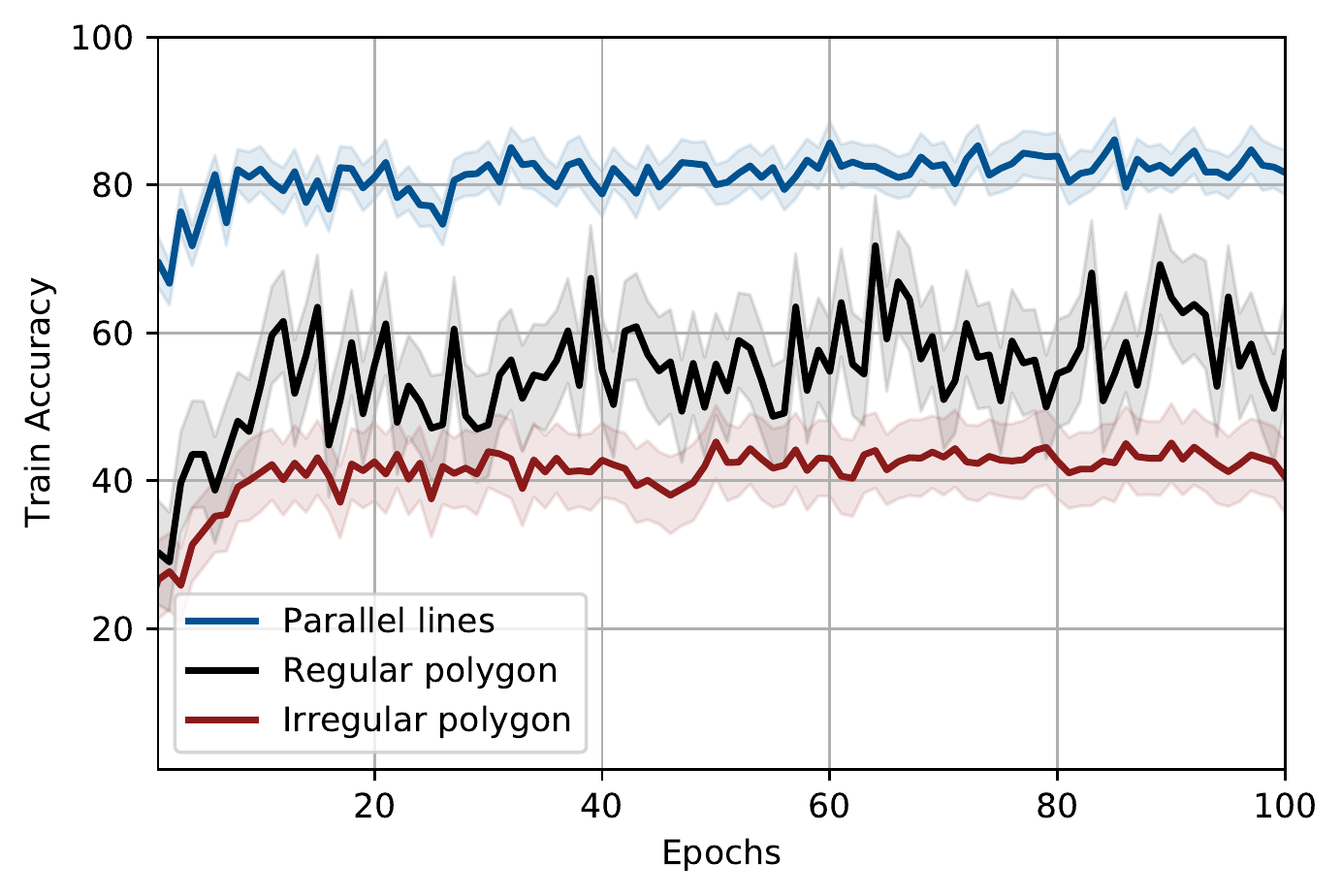}
        \caption{Accuracy on train data}
    \end{subfigure}%
    \begin{subfigure}[b]{0.5\textwidth}
        \centering
        \includegraphics[height=1.8in]{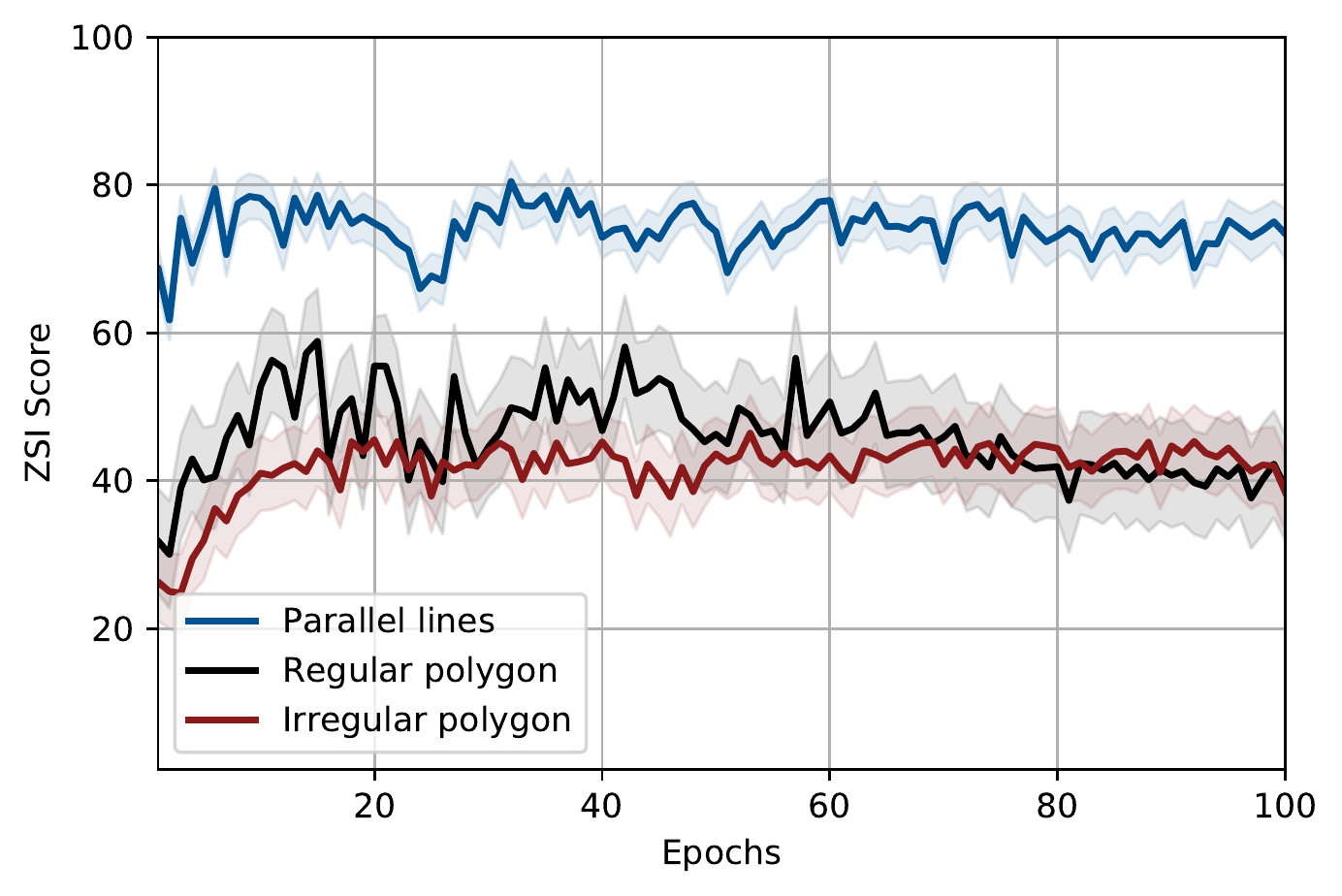}
        \caption{Zero-Shot Intelligence score ZSI on novel geometric shapes}
    \end{subfigure}
    \caption{Performance of Generative Adversarial text-to-image synthesis on the 3-9 world dataset. The results were averaged over several initializations of internal parameters. }
    \label{fig:gat}
\end{figure*}

\subsection{Computing the $\psi$ score}

Since deep neural networks continue to remain mostly as black-box function approximators \citep{chakraborty2017interpretability}, two models that produce the same results with partially correct zero-shot results, cannot be awarded the same score. For the 3-9 World dataset, let us consider that two algorithms on zero-shot testing generate four lines in the place of five lines. Though the answers are partially correct to the same extent, while one algorithm might have generated four lines using regression techniques, the other might have generated four lines through recursive attention methods. While it cannot be predicted which algorithm has reasoned on the task, we awarded a reduced score to both algorithms. Hence, partial scores were awarded in the range $25\ to\ 75$, while full scores were awarded only when the results are completely correct. Such partial scores were derived in proportion to the closeness of the generated image to the description. For example, for a text description of \textit{'five lines'}, if four lines were generated, a partial score proportional to 4/5 was awarded in the range of  $25\ to\ 75$. Complete correctness of the results were relaxed with permissible error limits, as derived from the experimental results from the top 5 performances of human test takers. $\psi$ score was calculated independently for parallel lines, irregular polygons and regular polygons.

\subsection{Rule-based evaluator}

A rule-based evaluator was used to perform automatized evaluation of every image for its closeness to the given textual description. The evaluator evaluates images during dataset generation, as well as to deduce the  $\psi$ score of the given text-to-image synthesis model. Douglas-Peucker algorithm \citep{douglas1973algorithms} was used to detect the contours of the generated figure. To accommodate minor variations, upon Gaussian blurring, canny edge detectors \citep{canny1987computational} were used for detecting lines. To identify the color of the figure, the k-means clustering \citep{hartigan1979algorithm} for pixel level RGB values were computed. Since Gaussian blurring is used, it is sufficient for the text-to-image synthesizer to generate a figure of the specified color either in the first dominant or second dominant region of the color cluster.

\section{Experiments and results}

\begin{figure*}[t!]
    \centering
    \begin{subfigure}[b]{0.5\textwidth}
        \centering
        \includegraphics[height=1.8in]{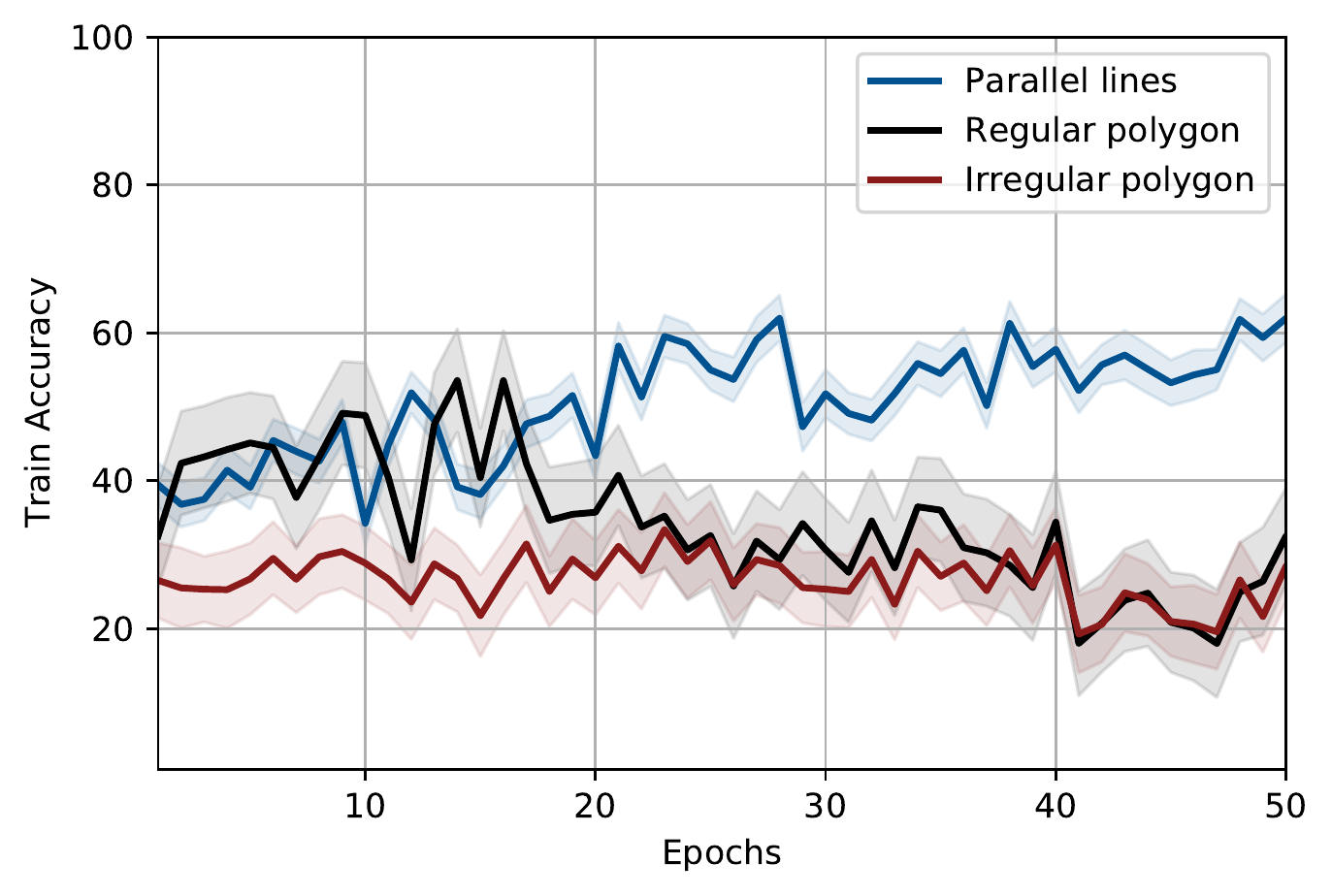}
        \caption{Accuracy on train data}
    \end{subfigure}%
    \begin{subfigure}[b]{0.5\textwidth}
        \centering
        \includegraphics[height=1.8in]{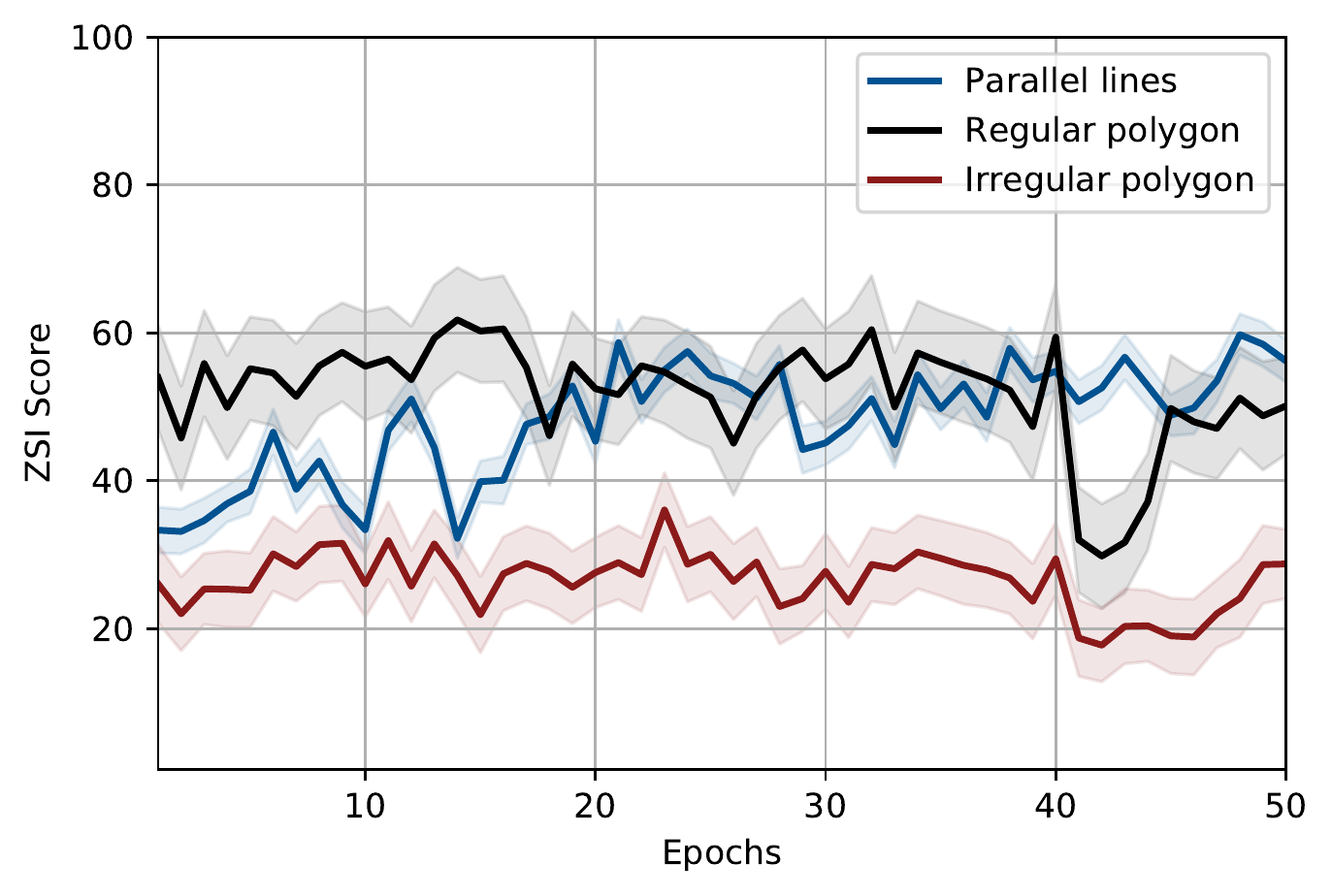}
        \caption{Zero-Shot Intelligence score ZSI on novel geometric shapes}
    \end{subfigure}
    \caption{Zero-Shot Intelligence performance of AttnGAN on the \textit{3 to 9 world} dataset. The results were averaged over several initializations of internal parameters.}
    \label{fig:att}
\end{figure*}

\subsection{Conditional generative networks: Generative adversarial text-to-image synthesis}

Generative Adversarial Text-to-Image Synthesis \citep{reed2016generative} is one of the widely used text augmented conditional generative network for text-to-image synthesis. The performance of this vanilla model on the \textit{3-9 World} dataset precisely  illustrates how the model fails to rapidly combine its learned representations for zero-shot generations. Interestingly, the model performs comparatively better in generating disconnected-parallel lines than while connecting such lines to form a meaningful polygon. More specifically, the model struggles in generating a irregular polygon than a regular polygon. This evinces that the convolutional and deconvolutional neural networks of the generative model develop specific filters for regular shapes but fail to map randomly varying irregular polygons. From time to time, the model generates completely connected polygons, illustrating its ability to generate connected shapes. From figure \ref{fig:gat}, the $\psi$ scores across epochs illustrates that the zero-shot performance is comparatively consistent but poor. Hence, a proactive-rapid optimizer can potentially boost the zero-shot performance of the model.

\subsubsection{Cosine distance between the sentence embeddings of skip-thought vectors for '3-9 World' dataset}

\begin{figure}[h!]
  \centering
  \includegraphics[width=0.5\textwidth]{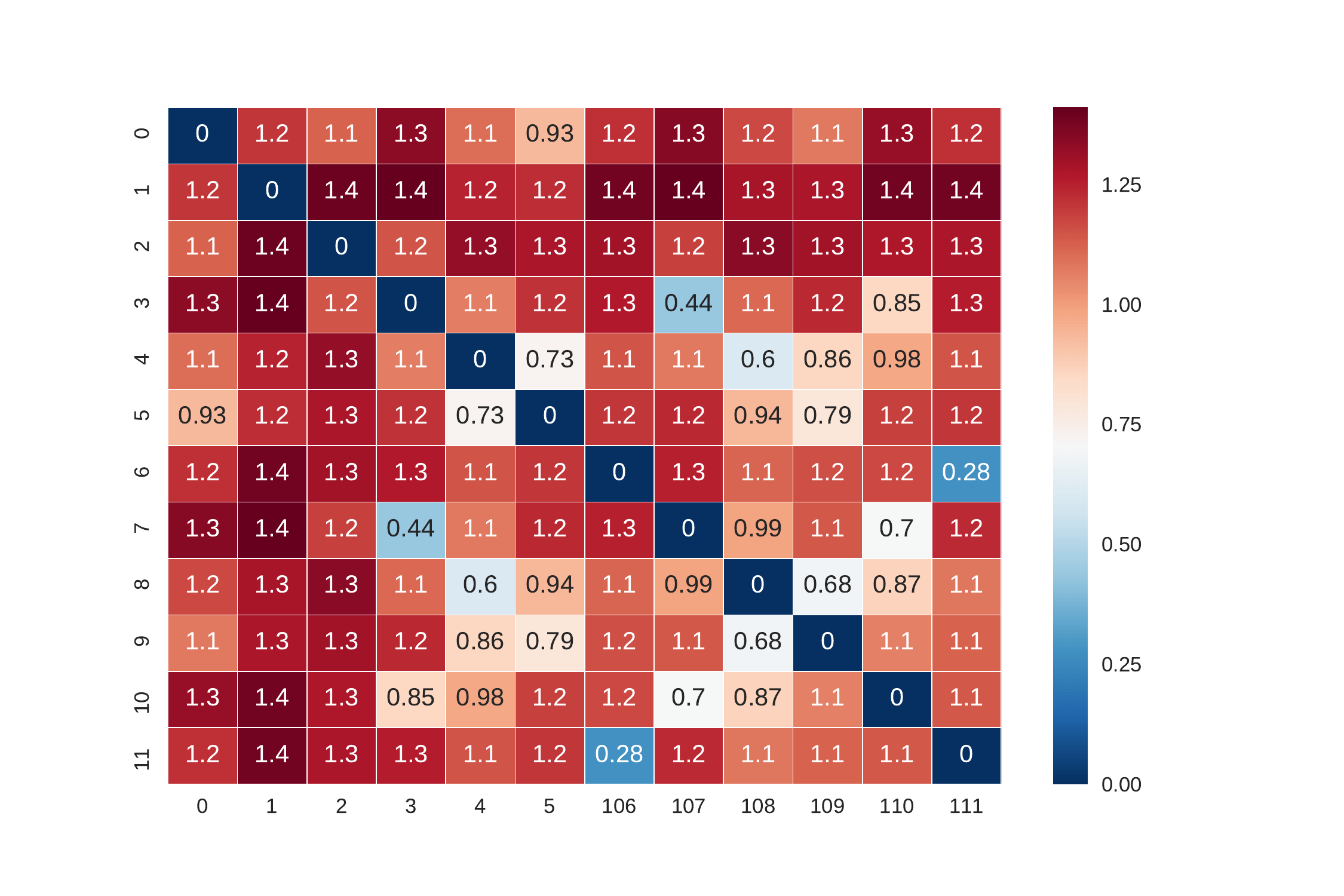}
  \caption{Cosine distance between the sentence embeddings of skip-thought vectors for '3-9 World' dataset. Columns $0\ to\ 5$ represent sentences of the same image while columns $106 \ to \ 111$ represent sentences of the same image from a different class. }
\end{figure}

The similarity between various sentence embeddings were computed. The cosine distance between the skip-thought vectors \citep{kiros2015skip} illustrates that all the skip thought vectors used to generate images were highly distinct and dissimilar. Upon training, the model established strong correlation between similar textual descriptions. For example, the text 'three green colored lines' and 'the image contains three lines that are green in color' generated similar images, as computed by the similarity between their $\psi$ scores.

\subsection{Recurrently attentive generative networks: AttnGAN}

\begin{figure}[h!]
  \centering
  \includegraphics[width=0.4\textwidth]{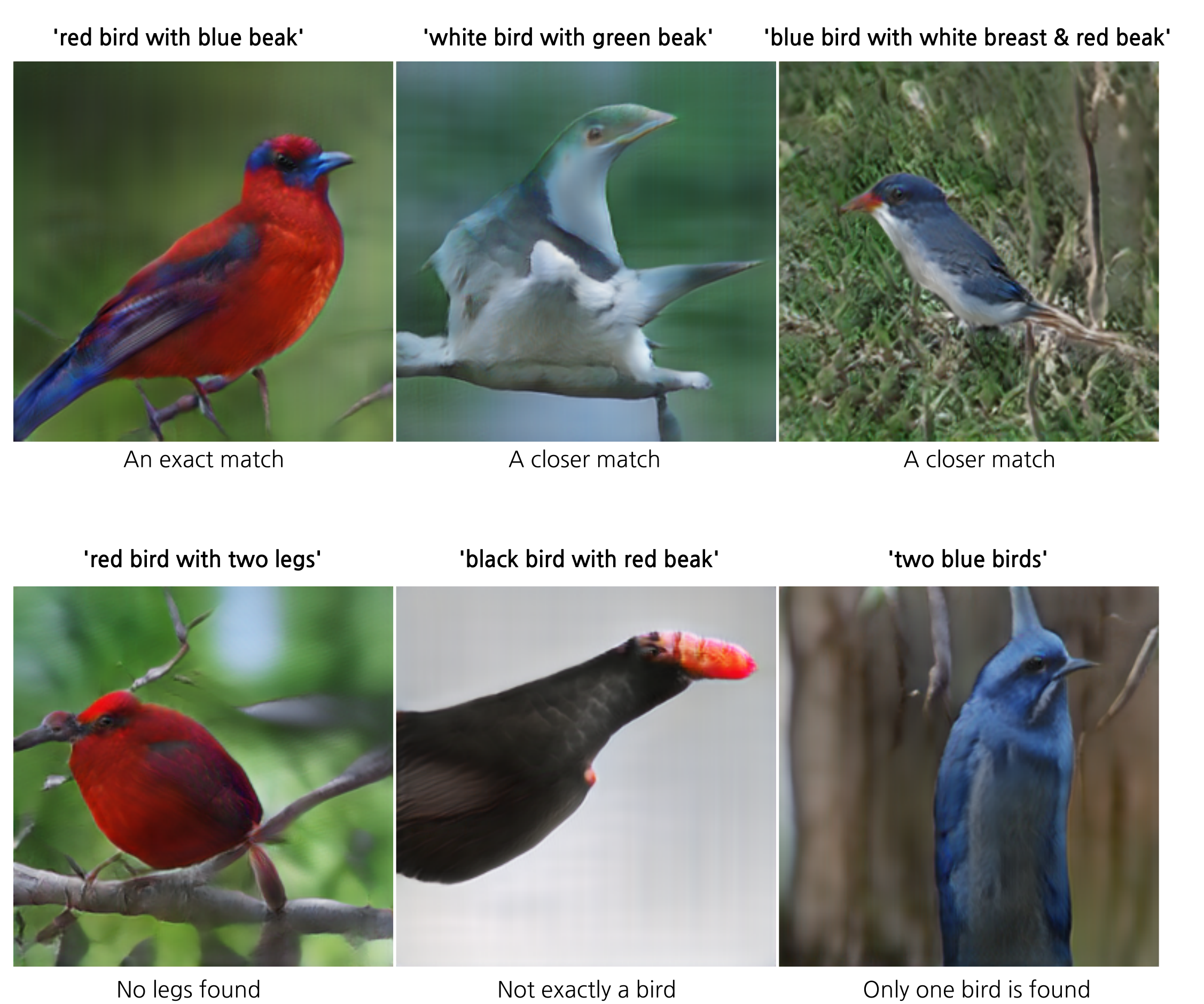}
    \caption{AttnGAN generated images when trained on CUB-2011 dataset. The images illustrates that the ability of the model to spatially and numerically reason are experimental. Note that the displayed textual inputs are not from completely unseen classes. }
        \label{fig:att2}
\end{figure}

We chose another state-of-the-art text-to-image generation model, Attentional Generative Adversarial Networks (AttnGAN) \citep{xu2017attngan}. In addition to the multiple generative networks stacked upon one another, its recurrent attention on word embeddings is an interesting feature of this model. As reported in their actual work, upon training on the CUB \citep{wah2011caltech} and COCO datasets \citep{lin2014microsoft}, the model out performed all state-of-the-art text-to-image generation methods. The best inception score on corresponding datasets being $4.36 \pm  .03$ and $25.89 \pm .47$ respectively. Figure \ref{fig:att2} illustrates the zero-shot generations by AttnGAN. Though the images are comparatively more realistic, the generated images do not match the input text by all means. Hence, an empirical research on the fundamental abilities of the model to spatially and numerically reason are essential.

We trained AttnGAN on the \textit{3-9 World} dataset over resized  $256 \times 256$  pixel images. The model was trained on various DAMSM encoders over the $200^{th}$ and $450^{th}$ epochs. Yet, the $\psi$ scores remained same. Notably, despite its success on fine grained image generation, the model was not able to demonstrate an improved $\psi$ score. From the $\psi$ scores in figure \ref{fig:att}, it is evident that the recurrent attention method has not improved the generalizing and few-shot learning abilities of the model. Interestingly, the vanilla Generative Adversarial Text-to-Image Synthesis model outperforms AttnGAN in the  $\psi$ scores. Such low $\psi$ scores and their tendency to fluctuate across epochs indicates the model's poor generalization ability and inconsistency respectively. Hence, the multiple recurrently attentive method and introduction of noise into AttnGAN are clearly subjected to further examination. 

\subsection{Analogous experiments on human beings}

We observe that the task of zero-shot text-to-image synthesis and the evaluation metric $\psi$ are not exactly applicable to the human test takers. Yet, we conducted analogous experiments on humans (n = 21, mean age = 27.3, s.d = 11.3), only to have an estimate of human-level performance. All test-takers were literate and at-least had a high school degree. The test takers were then asked to draw images of regular and irregular polygons of randomly chosen sides in the range $3$ to $30$. During the questionnaire, the human test-takers noted on average that they have rarely come across specific polygons that were more than 13 sided. Hence, sketches of polygons that were more than 13 sided can be considered analogous to zero-shot generation. The disadvantage of this approach being, no direct comparison could be drawn to the generative models' zero-shot performances on \textit{3 to 9 world} dataset. The scores were computed manually, partially comparable to the $\psi$ scoring metric.

\begin{figure}[h!]
  \centering
  \includegraphics[width=0.4\textwidth]{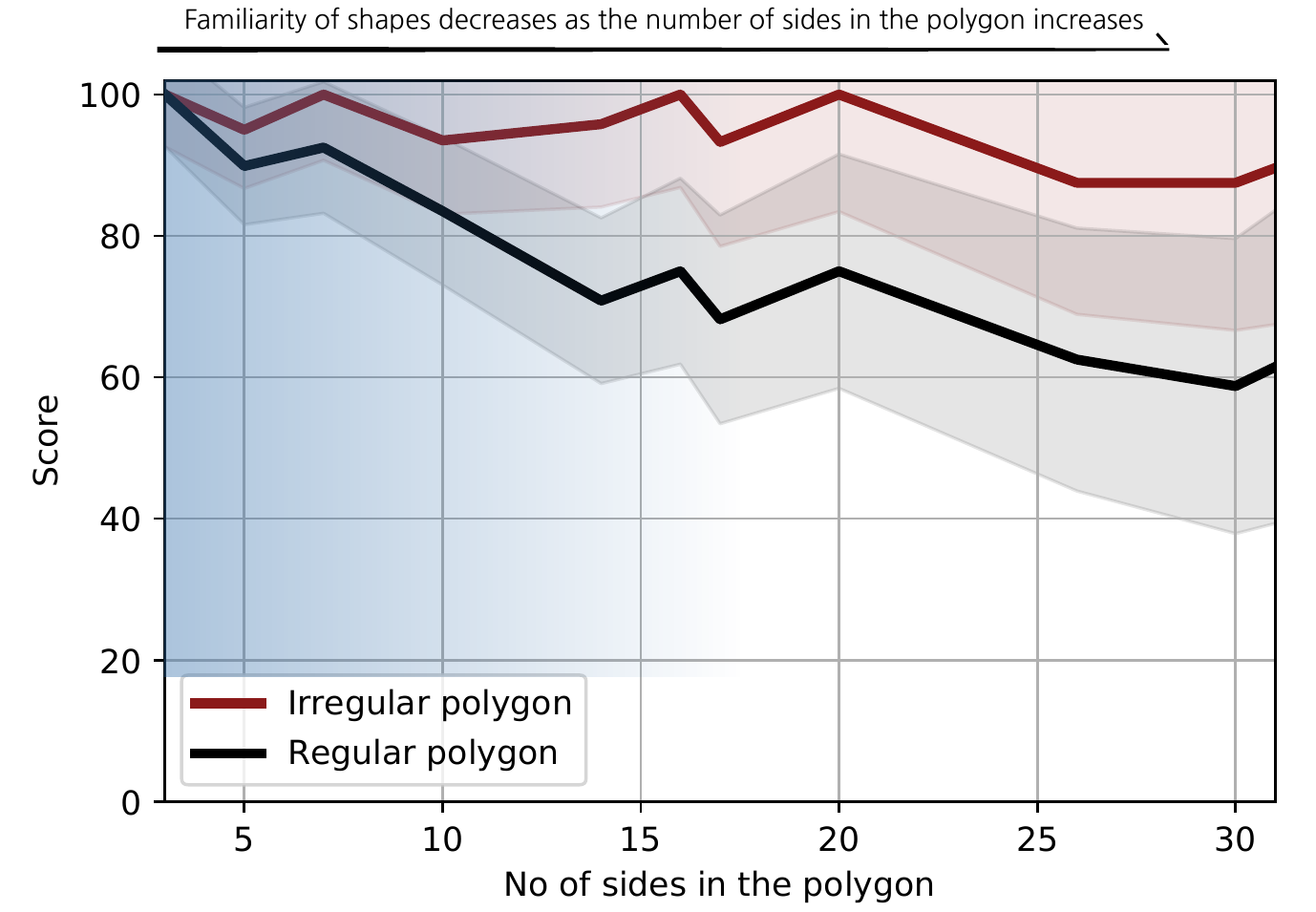}
    \caption{Analogous experiments on human beings. The peaks in-between are from even numbered polygons. The drop at around number five was owing to few participants skipping easier ones.}
\end{figure}

Multiple attempts by a human test-taker were ignored, until the person confirmed his final answer. Inspiration for such an evaluation technique is derived from the multiple attempts that were made during discovery of gravitational waves \citep{einstein1937gravitational}. Despite several editions and updates to his initial predictions, Einstein's ultimate prediction preceded the experimental confirmation. Hence, such multiple attempts can be neglected and the learning can still be considered as zero-shot learning.  

The analogous experiment results on humans show that the fundamental way in which humans and machine learning algorithms generate geometric figures differ greatly. While machine learning algorithms perform comparatively better on regular polygons via rote memorization of train images, human beings consistently generate irregular polygons with ease (until they were tired of sketching figures that were more than 30 sided).

\section{Discussion}

\subsection{Proactive Optimization for Zero-Shot Learning}

\begin{figure}[h]
  \centering
  \includegraphics[width=0.45\textwidth]{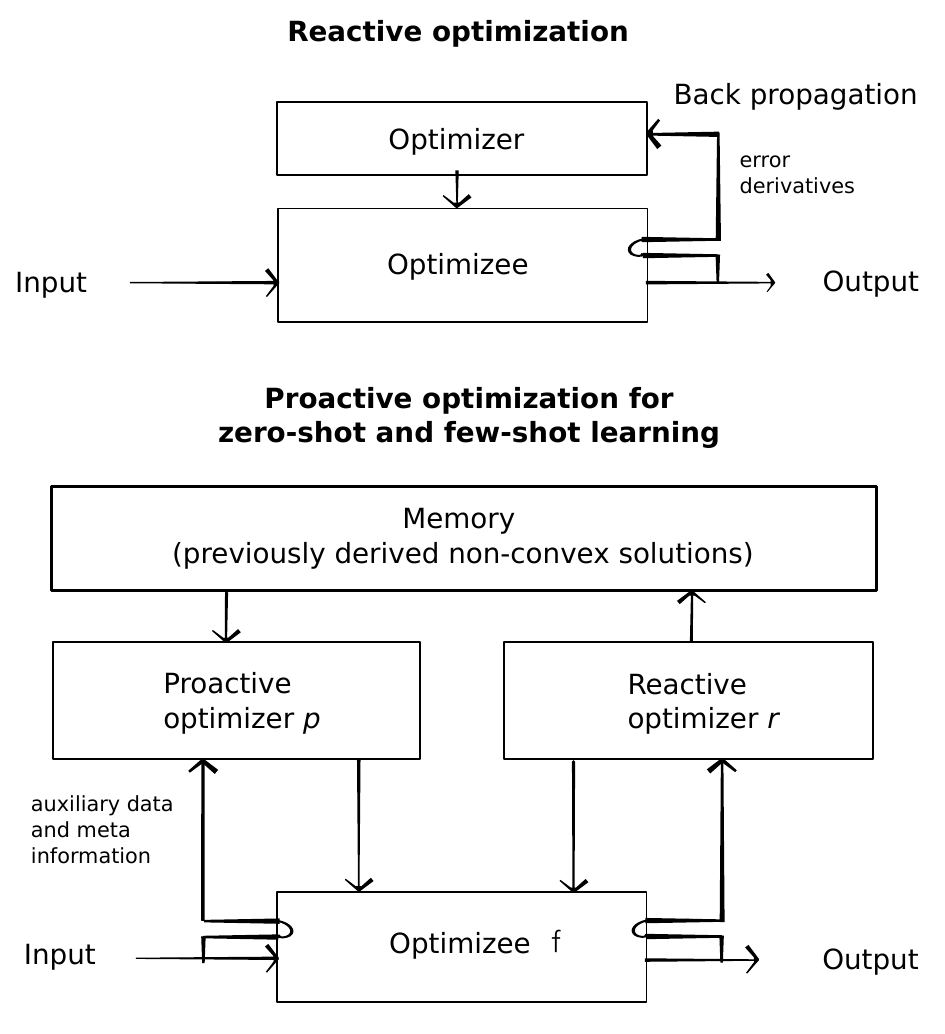}
    \caption{Reactive and proactive optimization}
\end{figure}

\begin{algorithm}[h]
   \caption{Proactive optimization}
   \label{alg:example}
\begin{algorithmic}
   \STATE {\bfseries Input:} Data $x_n$, explicit or implicit auxiliary information $a_n(x_i,y_i)$, meta-level information $m_i(f,x_i)$ available within the system $f$, reactively optimized parameters $W$, pro-actively optimized parameters $W^*$, loss function $L(y_i,f(x_i,W,W^*))$, proactive optimizer $p$, reactive optimizer $r$, zero-shot input $x_z$, zero-shot ground truth $y_z$:
   \REPEAT
   \STATE Initialize $W,\ W^*$ for $f:X\rightarrow Y$
   \FOR{$i=1$ {\bfseries to} $n$}
   \IF{$L(.)_{train\ set}> L(.)_{validation\ set}$} 
   \STATE $W_{t+1}^* = W_t^*+p( W_t^*,a_i(x_i,y_i), m_i(f,x_i))$ 
   \STATE $W_{t+1} = W_t+r( \nabla_\theta L(W_t))$
   \ENDIF
   \ENDFOR
   \UNTIL{$f(x_z,W,W^*)=y_z$}
\end{algorithmic}
\end{algorithm}

Non-convex optimization methods in deep neural networks have increasingly focused on skipping saddle points \citep{dauphin2014identifying, ge2015escaping,anandkumar2016efficient, bottou2016optimization} and local minima while minimizing a specific loss function $L$. As sparse data necessitates \citep{duchi2013estimation}, demand for few-shot learning and generalization of concepts across tasks are essential. Yet, the subject has been rarely focused. Here, we propose pro-active non-convex optimization methods for generalization and zero-shot learning. Given a training set $\{(x_n,y_n),n=1...N\}$,  optimizing the internal parameters $W$ of a system $f$ upon reception of the back-propagated gradients through supervised data $y_n$ is a reactive way of learning  $f:X\rightarrow Y$. Proactive methods of optimization includes optimization of internal parameters $W^*$ through explicit or implicit auxiliary data $a_z(x_z,y_z)$ about the unseen class $x_z$, by utilizing the the meta-level information $m_z(f,x_z)$ computed by the system $f$ and uniquely combining previously derived non-convex solutions, that were stored through external memory augmentation \citep{sprechmann2018memory}. Recent approaches \citep{romera2015embarrassingly, munkhdalai2017meta} to few-shot learning have focused on a fully or partially proactive methods for faster adaptation. The proactive approach to optimization aids in utilizing the auxiliary cues and meta-level information of the task, thereby aiding rapid learning. Both reactive and pro-active optimizers can exist in the same system. Such parallel streams of learning are analogous to the complementary learning systems in the human brain \citep{o2014complementary}.

\section{Conclusion}

In this paper, we proposed a new range of zero-shot learning tasks to evaluate the generative model's internal consistency and generalization abilities. Our empirical research work on state-of-the art text-to-image generation model exhibits a huge gap between human-level rapid learning and the few-shot methods in machine learning. Despite its excellence in fine grained image generation by the recurrently attentive generative adversarial network, the model's internal consistency is lower than the vanilla conditional generative adversarial networks. Hence, the performance of a model on the proposed 2d-geometric generalization tasks, can be used as an effective method to compute the reliability of a model. Here, the reliability measure indicates the ability of the model to consistently perform human-level spatial and numerical reasoning tasks across classes. The dataset \textit{Infinite World} will aid in the development of a new range of optimization methods that can rapidly learn through generalization.

\section*{Acknowledgments}  

The authors thank SAILING lab researchers, visitors at the Machine Learning Department - CMU, the Department of Cognitive \&  Developemental Psychology - TU Kaiserslautern, Roy Maxion, Marcus Liwicki, Francisca Rodriguez for their valuable reviews \& discussions.

\bibliography{example_paper}
\bibliographystyle{icml2018}


\appendix
\section{Appendix}

\subsection{Exception handling }

In the dataset, in addition to the illustrated ways of connecting lines to form a polygon, there can be many other ways in partially or completely forming a polygon. For example, a star polygon is also a regular polygon, but the \textit{Infinite World} dataset does not depict one. Hence, the rule based evaluator has been assigned to notify whenever it has potentially encountered such an exception. Though no such exceptions were generated in the succeeding experiments, any detected exceptions were assigned to be saved separately for manual evaluation. The selection of such exceptions were based on the unique geometric features of the exception. For example, an interestingly-connected open-ended figure would have more free edges than a irregular or regular polygon.

\begin{figure}[h!]
  \centering
  \includegraphics[width=0.20\textwidth]{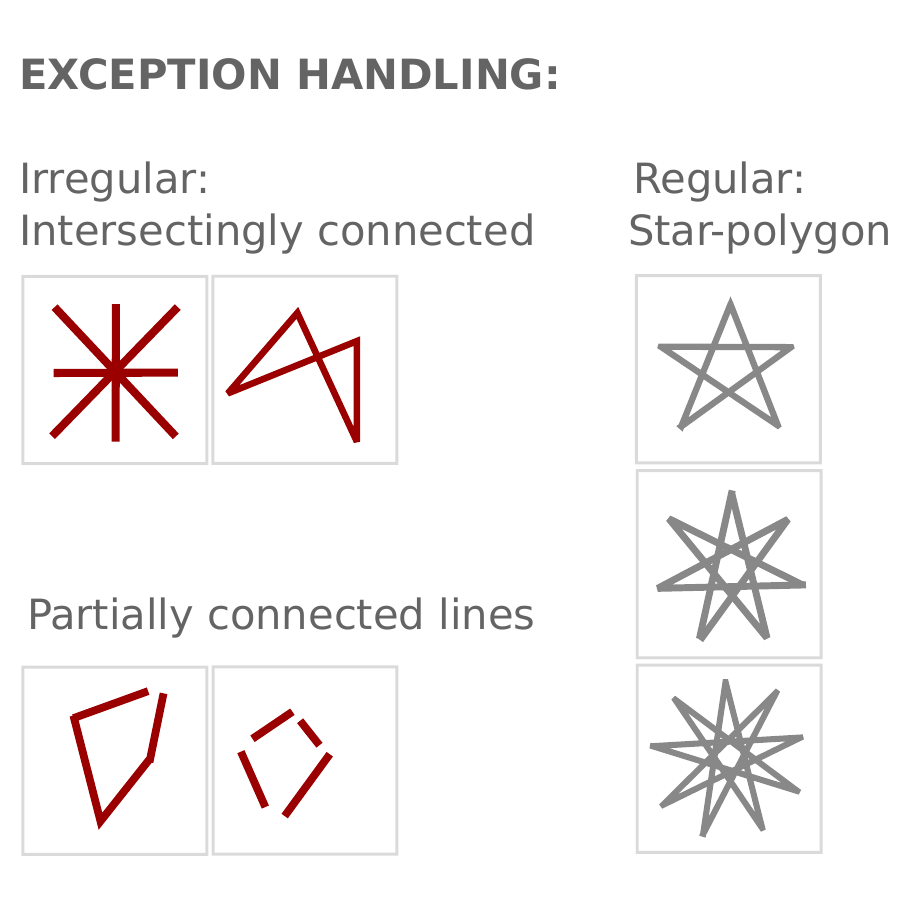}
    \caption{Handling of exceptional cases}
\end{figure}

\subsection{One-shot classification performance of Meta Networks, a partially pro-active optimizer}

Since the task of image classification is different from image generation, we performed few-shot classification on state-of-the-art one-shot classifier Meta Networks \citep{munkhdalai2017meta}. Meta Networks uses a partially proactive and as well as a reactive optimizer for few-shot image classification. Since the proactive optimizer is dependent on the error derivatives from a support-set, the proactive optimizer is not completely independent from the back propagation technique. The model performed one-shot learning at above $70\%$ accuracy (accuracy as defined in the original paper). Note that, only one-shot performance was measured and the ability to numerically reason beyond number $9$ was not computed in this experiment.

\begin{figure}[h]
  \centering
  \includegraphics[width=0.40\textwidth]{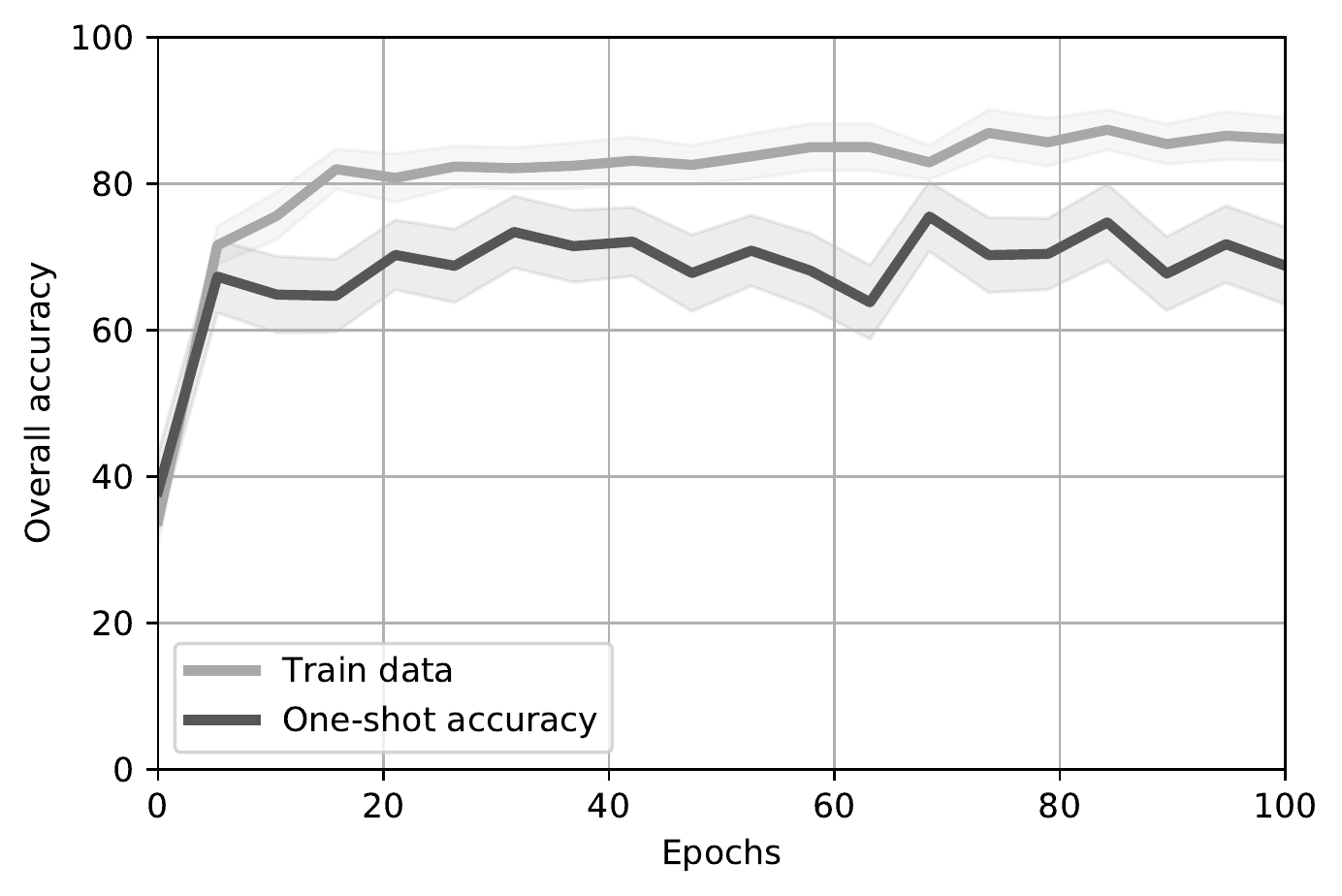}
    \caption{One-shot classification performance of Meta Networks, a partially proactive optimization based image classifier, on 3-9 World }
\end{figure}

\subsection{Applications and future work}

\begin{figure*}[t!]
  \centering
  \includegraphics[width=\textwidth]{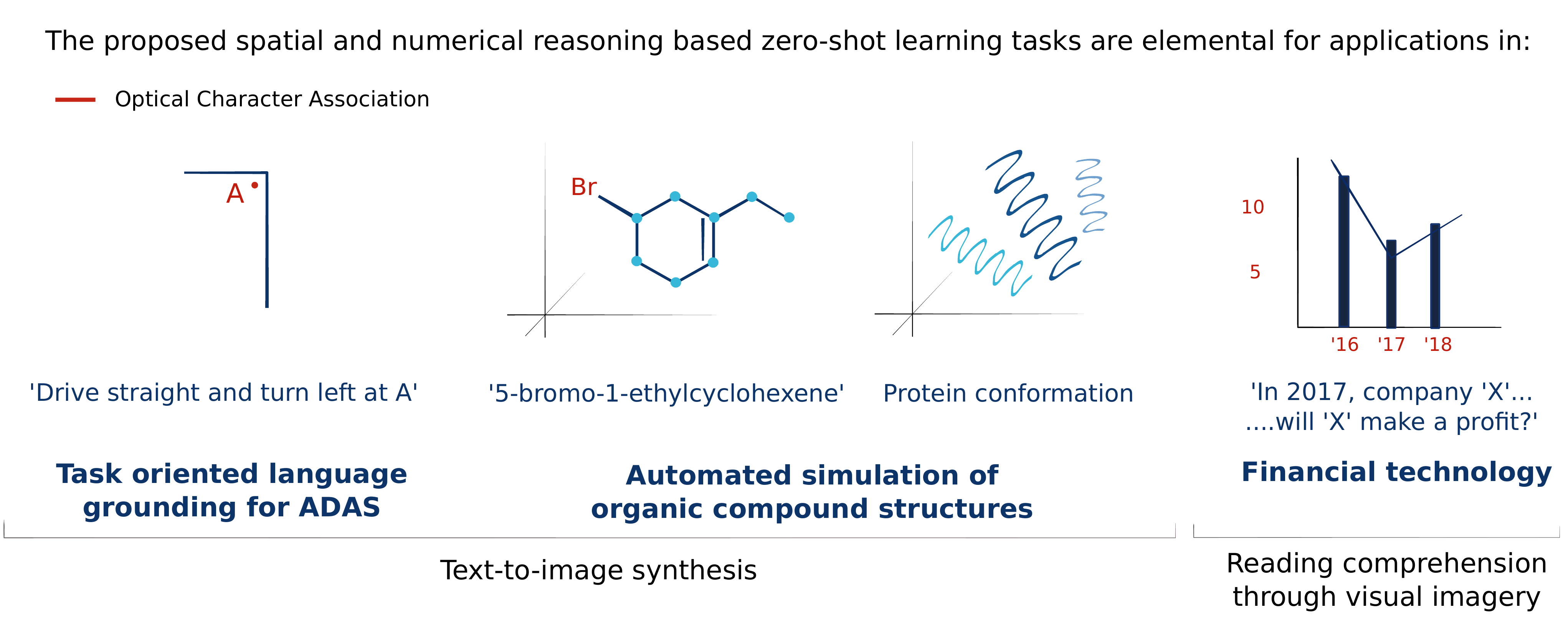}
    \caption{Few future applications of description based zero-shot learning. In addition to text-to-image synthesis and classification tasks, the dataset can also be extended to FigureQA \citep{kahou2017figureqa} type advanced VQA tasks}
\end{figure*}

To perform geometric generalization, it requires both numerical and spatial reasoning abilities. Hence, future applications of such zero-shot learning are enormous. The newly introduced tasks are elemental for applications in simulation based planning for autonomous vehicles, automated simulation of protein conformation, improving reading comprehension through image generation, intrinsic reward prediction for reinforcement learning through simulation of future states, etc., To enable such applications, the dataset can be further extended for more than two spatial dimensions and for optical character association to the nearest geometric features. Further scope of development lies in expanding the verbal and visual corpus to enable such models to acquire knowledge directly from text books.

\begin{figure*}[ht!]
  \centering
  \includegraphics[width=\textwidth]{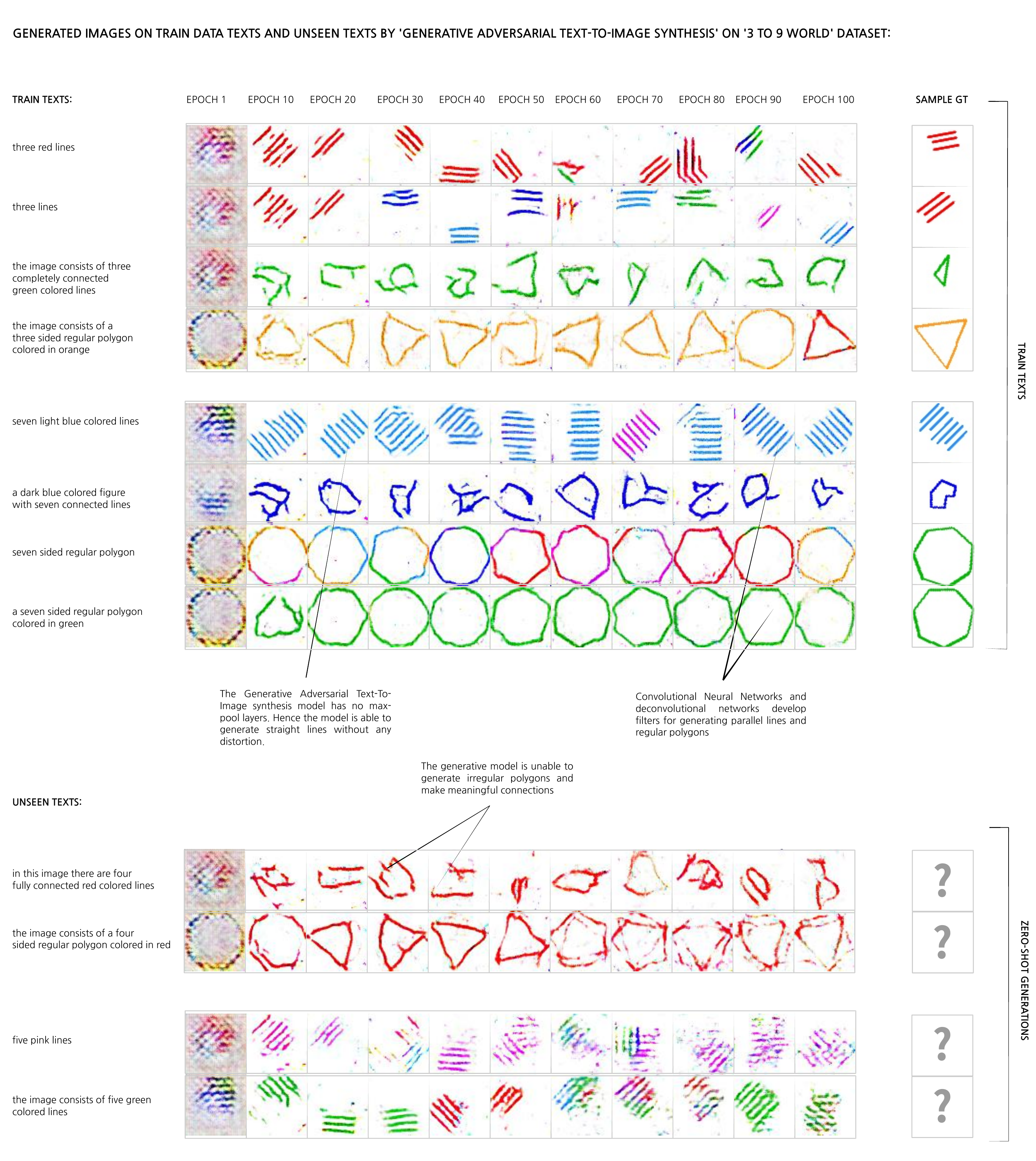}
    \caption{Generated images on train texts and unseen texts from Generative Adversarial Text-to-Image Synthesis on 3-9 World dataset. The displayed images were chosen in random.}
\end{figure*}

\begin{figure*}[ht!]
  \centering
  \includegraphics[width=\textwidth]{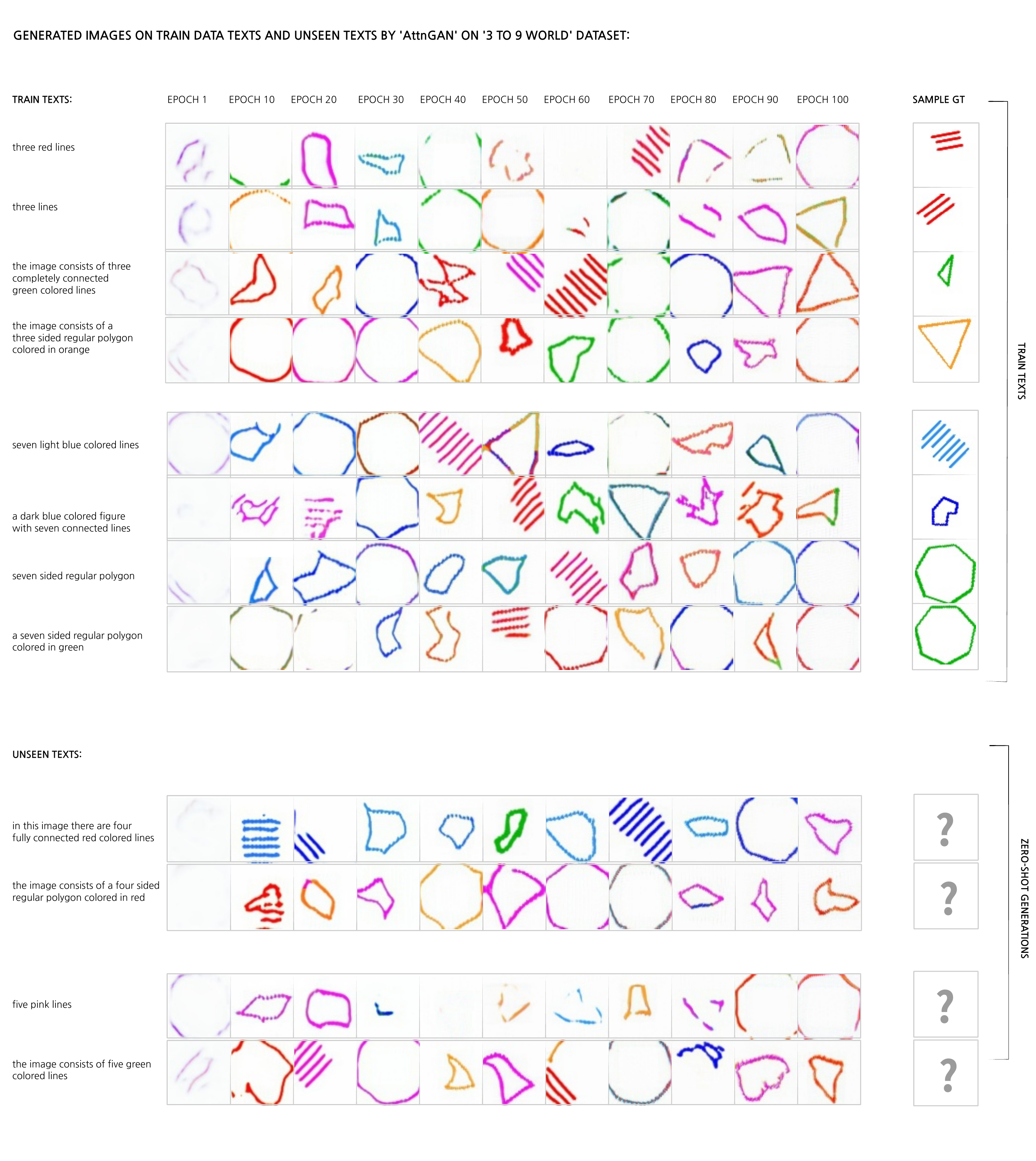}
    \caption{Generated images on train texts and unseen texts from AttnGAN on 3-9 World dataset. The displayed images were chosen in random.}
\end{figure*}

\begin{figure*}[ht!]
  \centering
  \includegraphics[width=\textwidth]{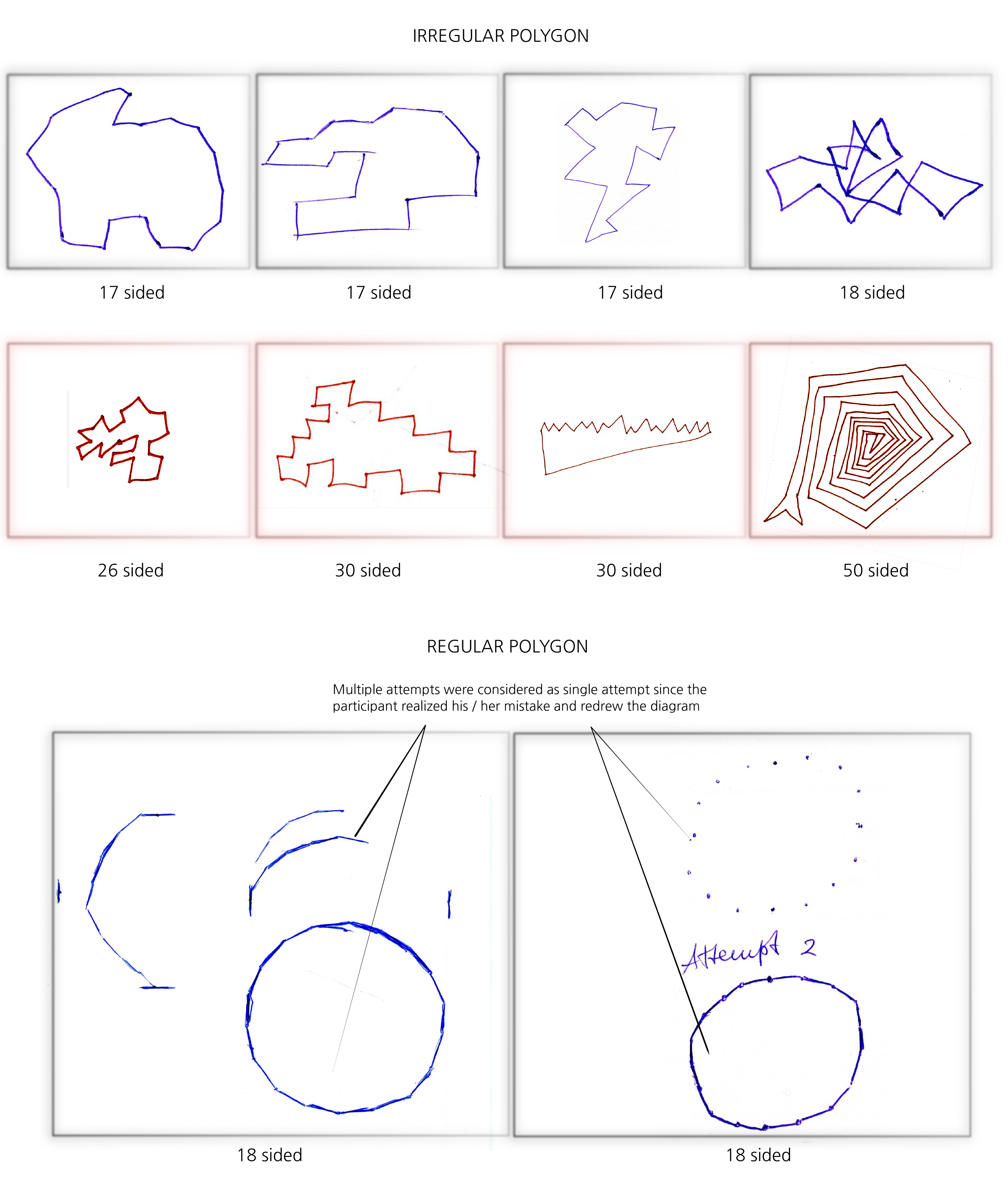}
    \caption{Analogous 2d-geometric generalization tests for human beings. All displayed results were awarded full scores.}
\end{figure*}

\pagebreak
\newpage

\end{document}